\pdfoutput=1
\documentclass[sigconf,nonacm,balance=false]{acmart}
\usepackage{amsmath}
\usepackage{booktabs}
\usepackage{graphicx}
\usepackage{placeins}
\makeatletter
\renewcommand\@subsubsecfont{\sffamily\upshape\bfseries}
\renewcommand\@parfont{\upshape\bfseries}
\makeatother
\usepackage{pifont}
\definecolor{markyes}{HTML}{1A9641}
\definecolor{markno}{HTML}{D7301F}
\newcommand{\cmark}{\textcolor{markyes}{\ding{51}}}
\newcommand{\xmark}{\textcolor{markno}{\ding{55}}}
\setcopyright{none}
\begin{document}
\title{Can Personal Requirements Become Supervision for Planning?}

\author{Junwei Wu}
\affiliation{%
  \institution{Emory University}
  \city{Atlanta}
  \state{Georgia}
  \country{USA}}

\author{Runze Yan}
\affiliation{%
  \institution{Emory University}
  \city{Atlanta}
  \state{Georgia}
  \country{USA}}

\author{Yuzhang Xie}
\affiliation{%
  \institution{Emory University}
  \city{Atlanta}
  \state{Georgia}
  \country{USA}}

\author{Hanqi Luo}
\affiliation{%
  \institution{Emory University}
  \city{Atlanta}
  \state{Georgia}
  \country{USA}}

\author{Ziwei Dong}
\affiliation{%
  \institution{Microsoft}
  \city{San Francisco}
  \state{California}
  \country{USA}}

\author{Jiaying Lu}
\affiliation{%
  \institution{Emory University}
  \city{Atlanta}
  \state{Georgia}
  \country{USA}}

\author{Xiao Hu}
\affiliation{%
  \institution{Emory University}
  \city{Atlanta}
  \state{Georgia}
  \country{USA}}

\author{Carl Yang}
\affiliation{%
  \institution{Emory University}
  \city{Atlanta}
  \state{Georgia}
  \country{USA}}

\renewcommand{\shortauthors}{Wu et al.}

\begin{abstract}
Managing multiple chronic conditions through diet requires satisfying disease- and medication-specific requirements jointly across meals and days while preserving established eating habits. These requirements are public in clinical guidelines and drug labels, yet turning them into a feasible plan for one person remains difficult for language-model agents. In our dietitian-approved weekly planning evaluation built from NHANES, thirteen models, from medically fine-tuned 8B models to GPT-6.1 Sol, pass fewer than 3\% of complete plans through direct generation. Explicit rules, higher reasoning effort, or model-directed tools leave GPT-6.1 Sol at no more than 8\%. We present Noura, which pairs LLM planning with executable personal specifications instantiated from public clinical guidance and individual health and dietary profiles. During planning, the specification grounds foods, optimizes portions, checks the complete multi-day plan, and guides revision. Our key insight is that a specification that rejects a plan can also determine what to learn from it. Noura turns executed plans and contextually credited revisions into supervision for compact models, recovering targets even from trajectories that never achieve complete success. The same specification supplies supervised targets, preferences, and rewards for post-training. On 4,993 held-out participants, Noura-SFT raises Qwen3-14B's complete-plan pass rate from 24.0\% to 72.5\% within the same execution loop, cuts rule violations by 97.5\% with no observed critical failures, and improves dietary quality. The locally hosted 14B planner incurs fewer rule violations than its GPT-6 Luna teacher. In controlled evaluation, it passes more plans than a 32B base model at 41\% of its amortized batch time. A matched comparison shows that executed targets reduce rule violations by 41\% over model-written proposals. Admitting improved plans from unsuccessful trajectories expands Qwen3-14B's candidate full-plan target pool 4.7-fold. Across four backbones from the Qwen, Phi, and Llama families, self-training alone reduces rule violations by 41--61\% on a 1,249-person population subset.

\end{abstract}
\keywords{Personalized planning, AI for Healthcare, LLM agents, learning from execution feedback, post-training, multimorbidity}
\maketitle
\raggedbottom
\setlength{\textfloatsep}{4pt plus 1pt minus 1pt}
\setlength{\floatsep}{4pt plus 1pt minus 1pt}
\setlength{\dbltextfloatsep}{4pt plus 1pt minus 1pt}
\setlength{\dblfloatsep}{4pt plus 1pt minus 1pt}
\setlength{\intextsep}{4pt plus 1pt minus 1pt}
\captionsetup[figure]{skip=2pt,belowskip=0pt}
\captionsetup[table]{skip=2pt,belowskip=0pt}
\section{Introduction}
\label{sec:introduction}

Chronic diseases impose a substantial burden of death and disability, and their frequent coexistence complicates long-term care. In 2023, 51.4\% of US adults reported two or more chronic conditions \cite{watson2025multiple}. Managing these conditions together requires coordinating disease-specific needs, medication-related restrictions, and everyday behaviors. Diet is a major modifiable factor in this burden: dietary risks accounted for an estimated 11 million deaths worldwide in 2017 \cite{gbd2019diet}. Dietary modification is foundational to cardiometabolic care, recommended from the outset and continued alongside medication when indicated \cite{jones2025hypertension,ada2026behavior}. Personalized dietary planning translates these clinical requirements into daily meals that jointly account for health conditions, medications, nutritional targets, and established eating habits across days and weeks.

Clinical guidelines and drug labels are publicly accessible on the Web, providing extensive guidance for dietary management \cite{ada2026behavior,nlm_dailymed,niddkckddiet}. Personalized agents using this Web guidance need user models that express individual health requirements and habits as jointly enforceable planning constraints across days. The United States has roughly one dietitian or nutritionist per 4,000 residents \cite{bls2026dietitians,census2026population}, while preparing a three-day meal plan took dietitians an average of 44 minutes in a feasibility study \cite{mundi2026mealplans}. In healthcare settings, sensitive patient information and institutional control requirements further motivate locally hosted models \cite{zhang2026onpremise}. Many hospitals lack the GPU resources needed to host large models locally, motivating compact models that fit existing infrastructure \cite{xu2026smallhospital}.

Large language model (LLM) agents can translate natural-language goals and heterogeneous information into personalized plans \cite{xie2024travelplanner}. However, their strength in open-ended generation does not ensure joint feasibility when decisions must satisfy hard constraints over a long horizon \cite{bui2026himap}. For dietary planning, this requires coordinating disease-specific nutritional targets and medication-related restrictions through food choices, portions, and their distribution over time. In our weekly planning evaluation, governed by a dietitian-approved specification, direct generation across thirteen general-purpose and medically fine-tuned models yields complete-plan pass rates below 3\%. Our tool-use comparisons likewise show that access to checking and portion-solving tools does not ensure complete-plan feasibility. More extensive reasoning and search can improve planning at additional inference cost, motivating learning from execution feedback to make more effective decisions within that budget \cite{singh2024restem,gehring2025rlef}. How can personal requirements guide both reliable planning and learning to plan more effectively under limited computational resources?

To address this question, we introduce \textbf{Noura}, a framework that combines LLM planning with \textbf{executable personal specifications} instantiated from online public clinical guidance and individual health profiles. The specifications connect food choices and quantitative execution to whole-plan checking and revision, coordinating disease- and medication-related requirements across days. Our key insight is that the same specification can determine what to learn from execution: \textbf{executed plans and revisions assessed within the complete plan} supply supervision even when a trajectory never achieves complete success. Noura thereby connects planning with post-training, drawing on both self-generated and stronger-model experience and using the specifications to construct supervised targets, preference comparisons, and reward signals.

We evaluate Noura at population scale and through controlled comparisons using public NHANES health and dietary records and FNDDS food identities and nutrient-composition data \cite{cdc_nhanes_dietary,usda_fndds}. We study open-weight and API-hosted planners and post-train Qwen3-8B, Qwen3-14B, Phi-4, and Llama-3.1-8B to support local hosting. Noura-SFT increases Qwen3-14B's complete-plan pass rate from 24.0\% to 72.5\% within the same execution loop at population scale, while improving HEI-2020 dietary quality. Under the same population execution protocol, the trained 14B planner incurs fewer rule violations than its GPT-6 Luna teacher. Matched serving comparisons also show higher complete-plan success in substantially less batch time than the base model with native reasoning enabled. A matched target-content comparison shows that executed plans provide more effective supervision for rule compliance and dietary quality than the corresponding model-written proposals. Learning exclusively from self-generated experience also reduces rule violations and generated output tokens across all three model families.

Our main contributions are:
\begin{itemize}
    \item \textbf{Problem formulation and systematic evaluation.} We formulate personalized dietary planning under interacting health, medication, temporal, and habit requirements, and establish a grounded evaluation protocol for complete-plan feasibility and personal specificity using public data and a dietitian-approved specification.
    \item \textbf{A unified framework for planning and post-training.} Noura uses executable personal specifications to guide planning and construct supervised targets, preference comparisons, and rewards, recovering supervision even from trajectories that never achieve complete success.
    \item \textbf{Reliable, efficient, and personalized planning.} Noura-SFT jointly improves complete-plan success, dietary quality, and batch inference efficiency at population scale. Controlled analyses establish the value of executed training targets, while complementary evaluations demonstrate recipient-specific plans, repeatable training gains, and effective self-training across model families.
\end{itemize}

\section{Related Work}
\label{sec:related_work}

\subsection{Personalized Dietary Intervention and Disease Management}
Trials support using individual physiological responses and health histories to personalize dietary care \cite{rein2022personalized,bermingham2024personalized}.
Making such requirements computable has a long line of work, from linear programs that meet public nutrient recommendations while staying close to observed intake \cite{maillot2010individual}, through rule-based meal planning for health conditions \cite{amiri2023flexible}, to medical agents that bring guidelines, medications, and longitudinal patient context into language-model reasoning \cite{wu2026nutriorion,lievin2026management}.
These strands address different parts of the problem. The quantitative systems neither plan with language models nor learn, while the agents reason over guidance without quantitatively verifying a complete multi-day plan. Neither compiles personal requirements into executable specifications that construct training targets.
Noura starts from the requirements themselves, compiling a person's profile and public guidance, observed intake, and cross-day consistency into one executable specification. Table~\ref{tab:related_work_positioning} positions it against these and the following systems along requirements, execution, and learning.

\begin{table}[t]
\centering
\captionsetup{textfont=normalfont}
\caption{Comparison of personal requirements, execution, and learning (Section~\ref{sec:related_work}). S/P/R denote supervised/preference/reward signals; Target indicates learning the executed plan.}
\label{tab:related_work_positioning}
\scriptsize
\setlength{\tabcolsep}{0.7pt}
\renewcommand{\arraystretch}{0.94}
\setlength{\aboverulesep}{0.25ex}\setlength{\belowrulesep}{0.45ex}
\begin{tabular*}{\columnwidth}{@{\extracolsep{\fill}}l@{\hspace{2pt}}|@{\hspace{2pt}}cccc@{\hspace{2pt}}|@{\hspace{2pt}}ccc@{\hspace{2pt}}|@{\hspace{2pt}}ccc@{}}
\toprule
\multicolumn{1}{l}{} & \multicolumn{4}{c}{\textbf{Personal requirements}} & \multicolumn{3}{c}{\textbf{Execution}} & \multicolumn{3}{c}{\textbf{Learning}} \\
\cmidrule(lr){2-5} \cmidrule(lr){6-8} \cmidrule(l){9-11}
\textbf{Method} & Profile & Guidance & Preferences & Horizon & Check & Solve & Revise & Signal & Self & Target \\
\midrule
Maillot et al.~\cite{maillot2010individual} & \xmark & \cmark & \cmark & \xmark & \cmark & \cmark & \xmark & -- & -- & -- \\
Amiri et al.~\cite{amiri2023flexible} & \cmark & \cmark & \cmark & \xmark & \cmark & \cmark & \xmark & -- & -- & -- \\
NutriOrion~\cite{wu2026nutriorion} & \cmark & \cmark & \cmark & \xmark & \cmark & \xmark & \xmark & -- & -- & -- \\
AMIE~\cite{lievin2026management} & \cmark & \cmark & \xmark & \xmark & \xmark & \xmark & \xmark & S, P & \xmark & \xmark \\
\midrule
LLM+P~\cite{liu2023llmp} & \xmark & \xmark & \xmark & \xmark & \cmark & \cmark & \xmark & -- & -- & -- \\
Hao et al.~\cite{hao2025formal} & \xmark & \xmark & \cmark & \cmark & \cmark & \cmark & \cmark & -- & -- & -- \\
CRITIC~\cite{gou2024critic} & \xmark & \xmark & \xmark & \xmark & \cmark & \xmark & \cmark & -- & -- & -- \\
HiMAP-Travel~\cite{bui2026himap} & \xmark & \xmark & \cmark & \cmark & \cmark & \xmark & \cmark & R & \cmark & \xmark \\
\midrule
Meerkat~\cite{kim2025meerkat} & \xmark & \xmark & \xmark & \xmark & \xmark & \xmark & \xmark & S & \xmark & \xmark \\
ReasonMed~\cite{sun2025reasonmed} & \xmark & \xmark & \xmark & \xmark & \xmark & \xmark & \cmark & S & \xmark & \xmark \\
HuatuoGPT-o1~\cite{chen2025huatuo} & \xmark & \xmark & \xmark & \xmark & \xmark & \xmark & \cmark & S, R & \cmark & \xmark \\
ether0~\cite{narayanan2025ether0} & \xmark & \xmark & \xmark & \xmark & \cmark & \xmark & \xmark & S, R & \cmark & \xmark \\
\midrule
ReST-EM~\cite{singh2024restem} & \xmark & \xmark & \xmark & \xmark & \cmark & \xmark & \xmark & S & \cmark & \xmark \\
ETO~\cite{song2024eto} & \xmark & \xmark & \xmark & \xmark & \cmark & \xmark & \xmark & S, P & \cmark & \xmark \\
NAT~\cite{wang2025nat} & \xmark & \xmark & \xmark & \xmark & \cmark & \xmark & \xmark & S & \xmark & \xmark \\
NExT~\cite{ni2024next} & \xmark & \xmark & \xmark & \xmark & \cmark & \xmark & \cmark & S & \cmark & \xmark \\
LeDex~\cite{jiang2024ledex} & \xmark & \xmark & \xmark & \xmark & \cmark & \xmark & \cmark & S, R & \cmark & \xmark \\
RLEF~\cite{gehring2025rlef} & \xmark & \xmark & \xmark & \xmark & \cmark & \xmark & \cmark & R & \cmark & \xmark \\
SIRL~\cite{chen2025sirl} & \xmark & \xmark & \xmark & \xmark & \cmark & \cmark & \xmark & R & \cmark & \xmark \\
SOLID~\cite{zhu2026solid} & \xmark & \xmark & \xmark & \xmark & \cmark & \cmark & \xmark & R & \cmark & \xmark \\
\midrule
\textbf{Noura (ours)} & \cmark & \cmark & \cmark & \cmark & \cmark & \cmark & \cmark & S, P, R & \cmark & \cmark \\
\bottomrule
\end{tabular*}
\end{table}

\subsection{LLM Planning with Executable Constraints}
Executing such a specification is a planning problem that language models handle poorly on their own. Agents violate constraints when assembling complete plans from retrieved information \cite{xie2024travelplanner}.
Classical planners and solvers handle search \cite{liu2023llmp,hao2025formal}, while verifiers steer candidate revision \cite{kambhampati2024modulo}. Tool feedback drives iterative correction \cite{gou2024critic}, and execution feedback refines whole trajectories under a monotonic acceptance rule \cite{zhang2026pivot}. Hierarchical schemes coordinate day-level planners under a global monitor while training a shared policy by reinforcement learning \cite{bui2026himap}.
These approaches coordinate generation with explicit constraint handling. Noura adopts this division of responsibility and turns grounded food choices and solved portions into supervised targets.

\subsection{Learning from Domain Knowledge and Execution Feedback}
How execution results can become supervision is the question of a third line of work.
Domain post-training builds reasoning data from textbooks and multi-model generation \cite{kim2025meerkat,sun2025reasonmed} or lets domain verifiers guide data construction and reinforcement learning \cite{chen2025huatuo,narayanan2025ether0}.
Learning from generated experience builds on several approaches. Expert Iteration trains a network to imitate the decisions that search improved \cite{anthony2017expert}, Dreamer learns from imagined trajectories \cite{hafner2019dreamer}, and ReST-EM filters sampled solutions with feedback before supervised updates \cite{singh2024restem}.
Agent tuning also learns from failure, contrasting failed rollouts with expert trajectories through preference optimization or fine-tuning on failed trajectories under a negative prompt \cite{song2024eto,wang2025nat}.
Constraint verifiers can likewise rank sampled responses into preference labels \cite{wang2025act}.
Execution-aware methods supervise test-validated repairs \cite{ni2024next,jiang2024ledex}, train multi-attempt use of execution feedback \cite{gehring2025rlef}, and use solver artifacts as rewards or as privileged context for self-distillation \cite{chen2025sirl,zhu2026solid}.
These execution-based methods use feedback to condition model-generated repairs, select targets, assign preferences or rewards, and re-score responses.
Noura applies the Expert Iteration principle to planning under a personal specification. Its executor grounds proposed foods and solves their portions, and the resulting executed plan becomes the target.
Because the specification both checks and improves each attempt, the planner can learn from its own attempts alone.
Revisions receive credit within the complete plan. A trajectory that never passes completely can still supply improved plans and credited revisions as targets. The same specification supplies supervised, preference, and reward signals, turning personal requirements into supervision.

\section{Method}
\label{sec:method}

Noura connects personalized plan execution to the learning of a more effective planning policy. It instantiates an executable specification from structured clinical and nutritional knowledge and an individual's health and dietary context. That specification coordinates clinical requirements, intake references, dietary plausibility, habit continuity, and constraints across days. During planning, it governs food grounding, portion optimization, evaluation, and revision. During learning, it converts the resulting decisions into supervision: the model learns to propose complete plans and contextual revisions with the food choices and quantities achieved through execution.

\begin{figure*}[t]
\centering
\includegraphics[width=\textwidth]{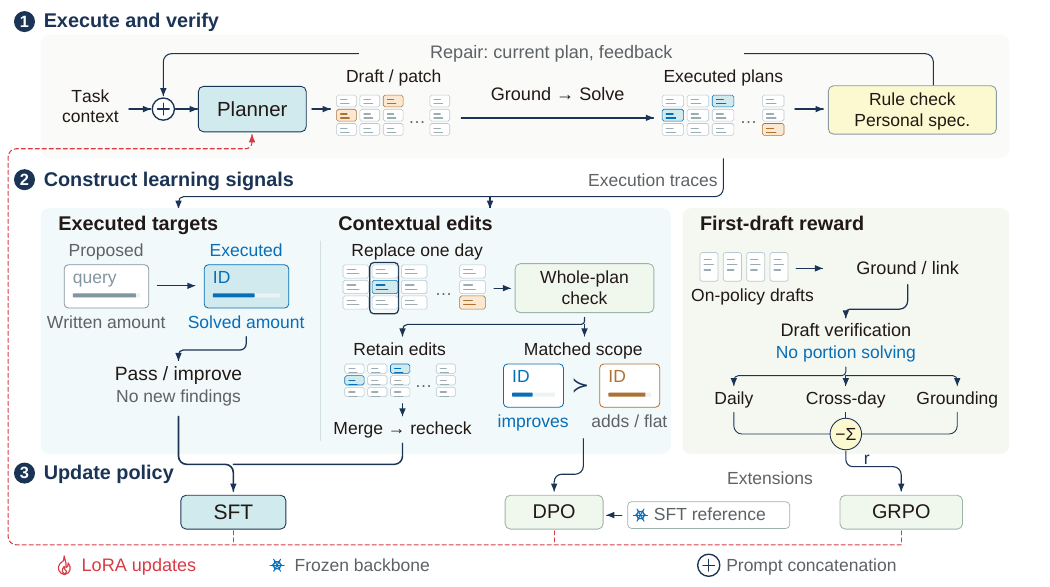}
\caption{Noura connects execution to learning: (1) execute and revise plans under a personal specification; (2) construct targets, preferences, and rewards; (3) update the planning policy. Solid arrows show computation or data flow, the dashed return shows LoRA updates, and the circled plus denotes prompt concatenation. Construction versions and checkpoint lineage appear in Appendix~\ref{app:method_details}.}
\Description{A three-stage diagram shows planning under a personal specification, construction of executed targets, preferences, and rewards from trajectories, and LoRA policy updates that feed back into planning.}
\label{fig:noura_method}
\end{figure*}

The shared specification provides three forms of learning signal (Figure~\ref{fig:noura_method}). Executed plans and credited revisions provide supervised targets. Preferences compare executed alternatives, and rewards evaluate fresh drafts. Thus, Noura learns from the outcomes of its own proposals as well as stronger-model trajectories, using the same personal requirements to determine what should be learned. The trained policy continues to operate with the execution and checking interface. We describe the specification, its execution loop, target construction, and post-training objectives in turn.

\subsection{Problem Definition}
\label{sec:method:1}

A personalized planning task has input \(x=(p,O,H)\): health context \(p\), observed dietary records \(O\), and a horizon of \(H\) days. Health context includes demographics, conditions, medications, anthropometrics, and available laboratory measurements. Dietary records contain foods, quantities, and eating occasions. Health context determines applicable requirements, while observed intake supplies quantitative references and eating habits.

A language-model policy \(\pi_\theta\) proposes food-and-portion plans for \(x\). We denote the grounded proposal by \(Z\) and its executed state by \(X=E_x(Z)\), where \(E_x\) adjusts portions under the personal specification \(\mathcal S_x\). Both organize food identifiers and gram quantities by day and meal: \(Z\) retains proposed portions, and \(X\) records executed portions. The policy revises proposals using execution feedback, and the system returns the retained plan's executed state.

\paragraph{Clinical rules and personal references} The clinical component of \(\mathcal S_x\) instantiates rules from five clinical guidelines and FDA labels for 20 drug products, covering 31 medications, four conditions, and two laboratory measures. Each rule records applicability, required inputs, individualized bounds or exclusions, severity, and provenance. Table~\ref{tab:scored_requirements} lists the scored requirements. Mean recorded intake supplies separate references for daily and weekly energy and, under warfarin, vitamin-K consistency~\cite{nlm_dailymed}. Following individual diet modeling~\cite{maillot2010individual}, this keeps energy near usual intake while changing dietary composition. A clinician-set energy target can replace the reference~\cite{jensen2014aha}.

\paragraph{Dietary structure and continuity} Operational planning criteria complement the clinical rules: daily checks constrain macronutrient balance, food concentration, and item counts, while cross-day checks constrain repetition, overlap, variety, and intake stability. Habit continuity requires a minimum energy share from observed foods or same-category substitutes, while allowing the full food catalog. An independent registered dietitian reviewed and approved the clinical rules and operational thresholds (Appendix~\ref{app:specification_review}).

\paragraph{Success and learning objective} Complete-plan success requires passing all evaluable clinical, plausibility, cross-day, and grounding checks on the food choices and executed quantities. Requirements needing unavailable inputs or clinical resolution are recorded separately. We learn a policy that reaches passing plans efficiently, using execution under \(\mathcal S_x\) to construct supervision, and assess dietary quality separately with HEI-2020. Appendix~\ref{app:method_details} gives thresholds and conflict handling.

\subsection{Executing and revising dietary plans}
\label{sec:method:2}

The execution loop converts a food-level proposal into a quantitatively evaluated plan and uses the personal specification to direct subsequent revisions.

\paragraph{From food proposals to quantitative plans} We ground proposed foods in the USDA Food and Nutrient Database for Dietary Studies \cite{usda_fndds}, checking supplied identifiers against their descriptions and resolving ambiguous entries through catalog candidates. Grounding establishes the nutrient composition needed to evaluate quantities in the proposal \(Z\) and its executed state \(X\). For food identifier \(c_{dmi}\) and executed grams \(g_{dmi}\), nutrient intake on day \(d\) is
\begin{equation}
\label{eq:method:v3:1}
\mathbf n_d=\sum_{m,i}g_{dmi}\,\mathbf a(c_{dmi}),
\end{equation}
where \(\mathbf a(c)\) is the catalog nutrient vector per gram.

For each day, a linear program adjusts portions within food-specific bounds, balancing deviation from the proposed quantities against severity-weighted constraint residuals. Forbidden food classes receive zero quantity bounds. The evaluator then checks the executed intake and assesses the complete plan across days. Food repetition, variety, and day overlap are computed from the grounded proposal. Nutrient intake and the energy share contributed by habitual foods use executed quantities. Keeping both representations makes the relationship between food choices, quantitative execution, and feedback explicit.

\paragraph{Revision under a shared specification} Feedback identifies failed checks, relevant quantities and thresholds, nutrient contributors, and grounding problems. Candidates are ranked lexicographically by
\begin{equation}
\label{eq:method:v3:2}
\begin{aligned}
\mathbf q_x(Z,X)=\bigl(&v_{\mathrm{critical}},v_{\mathrm{major}},
v_{\mathrm{moderate}},v_{\mathrm{minor}},\\
&v_{\mathrm{plaus}},v_{\mathrm{cross}},v_{\mathrm{ground}}\bigr),
\end{aligned}
\end{equation}
where the entries count clinical rule failures by severity, daily plausibility findings, cross-day findings, and ungrounded items. An all-zero vector denotes complete-plan success.

The planner receives the preceding proposal and its execution feedback and generates candidate meal replacements. Each candidate is merged into that proposal, grounded, executed, and evaluated under the same specification. The best candidate advances the rollout. The lowest-scoring retained state across all rounds determines the returned plan, with earlier rounds preferred on ties. Planning stops when the current state passes or the revision budget is exhausted. When collecting training trajectories, we record every candidate, its generating context, and its execution results. These records provide supervision for both producing a plan and revising one.

\subsection{Learning targets from execution}
\label{sec:method:3}

The learning object is a planning decision conditioned on the person and, for revision, the surrounding plan. Execution transforms a proposed decision into a quantitative outcome, while the specification determines whether that outcome supplies an admissible target. Full-plan targets pair the original request with the best admitted executed plan. Revision targets pair a recorded planning state and its feedback with executed changes credited in that context. Food identifiers make these targets computable, while food selection, quantities, and the scope of the changes carry the planning supervision. Replaying recorded trajectories constructs these targets without additional language-model generations or expert-written reference plans.

\paragraph{Executed targets for planning and revision} Let \(Z^0\) be an initial proposal and \(Z^\star\) the best retained proposal in its trajectory. For an admitted initial-planning example, the input is the original planning context \(u^0\), and the target serializes the executed plan:
\begin{equation}
\label{eq:method:v3:4}
(u^0,y^0),\qquad
y^0=\operatorname{Serialize}\!\left(E_x(Z^\star)\right).
\end{equation}
The target contains grounded food identities and executed portions, rounded to whole grams, with positive rounded quantities retained. This pairing makes the outcome of quantitative execution and iterative revision directly available as supervision for initial generation. Revision examples use the same executed representation for changed meals, paired with the recorded context that generated the candidate. Their content is selected by the contextual credit test below.

\paragraph{Target admission} Let \(F_x(Z)\) be a counter of findings obtained by evaluating \((Z,E_x(Z))\). Each coordinate identifies a failed check and its context: a rule or plausibility check on a day, a cross-day check with its associated food or day pair, or an ungrounded item. For counters \(\mathbf a\) and \(\mathbf b\), define
\begin{equation}
\label{eq:method:v3:3}
\mathbf a\prec\mathbf b
\quad\Longleftrightarrow\quad
\bigl(\forall k,\ a_k\le b_k\bigr)
\ \land\
\|\mathbf a\|_1<\|\mathbf b\|_1.
\end{equation}
This relation requires fewer findings overall and no increase in any finding coordinate, preserving the identity of the affected requirements. We admit an initial-planning example when the trajectory achieves complete-plan success, or when \(Z^\star\) occurs in a later round and \(F_x(Z^\star)\prec F_x(Z^0)\). The latter condition makes strictly improved executed plans available for learning before complete feasibility is reached.

\paragraph{Contextual credit for revision targets} A candidate can modify several days, whose contributions must be assessed in the surrounding plan. Let \(B\) be the preceding retained proposal, \(Q_j\) a candidate, and \(D_j\) its changed days. Write \(B\oplus_S Q_j\) for the plan formed by inserting candidate days \(S\) into \(B\). A construction evaluator \(C_{x,v}(B,Q)\) assembles executed daily states and recomputes cross-day and grounding findings under construction version \(v\). Candidate \(j\) receives credit on days
\begin{equation}
\label{eq:method:v3:5}
S_j=\bigl\{d\in D_j:
C_{x,v}(B,B\oplus_{\{d\}}Q_j)\prec F_x(B)\bigr\}.
\end{equation}
Each test retains the other days as context. Candidates with \(S_j=\varnothing\) are discarded. We combine the credited days and require \(C_{x,v}(B,B\oplus_{S_j}Q_j)\prec F_x(B)\) again to assess their joint effect. Among eligible candidates, we select the one with the greatest reduction in total findings, breaking ties by candidate order. Its target contains the changed meals on credited days, serialized with executed food identities and portions. Its input is the recorded context that generated that candidate.

Suppose a food with a four-day repetition limit already appears on three days. Two proposed replacements each resolve a local violation by introducing it on another day, without adding other findings. Individually, each can receive credit. Together, they introduce a fifth appearance. Rechecking the combined edits prevents these individually useful changes from becoming a target that adds a weekly violation.

The construction version \(v\) is fixed for each training corpus. Initial-plan admission and inference use the full finding counter. Appendix~\ref{sec:method_details:4} specifies the candidate-side checks in each version, the corresponding checkpoints, and the controlled retraining analysis.

Admission therefore operates on learning targets within a trajectory. An unsuccessful trajectory can contribute an improved executed plan, and a candidate that did not advance the rollout can contribute credited revisions. Together, these targets teach the model to anticipate quantitative execution and to make changes that improve a plan in its personal context. Their admission is determined by the specification, independently of which model generated the trajectory.

\subsection{Post-training compact planners}
\label{sec:method:4}

Executed-target SFT learns the planning policy from specification-assessed decisions. The same specification supports further learning through preferences over contextual revisions and rewards for fresh plans. We use standard SFT, DPO, and GRPO objectives. The task-specific construction lies in the targets, comparisons, and rewards supplied by execution.

\paragraph{Supervised planning and revision} We combine admitted full-plan and revision examples into \(\mathcal D\). For recorded context \(u\) and constructed completion \(y\), Noura-SFT optimizes
\begin{equation}
\label{eq:method:v3:6}
\mathcal L_{\mathrm{SFT}}(\theta)=
-\sum_{(u,y)\in\mathcal D}\sum_{t=1}^{|y|}
\log\pi_\theta(y_t\mid u,y_{<t}).
\end{equation}
The loss masks the input context and covers all completion tokens. Noura constructs supervision from planner-generated experience through the execution-target procedure above. Noura-SFT (self) uses trajectories generated by the corresponding base planner. The primary Qwen3-14B Noura-SFT configuration combines Qwen3-14B and GPT-6 Luna trajectories collected for the same training participants. Both sources undergo the same execution, target admission, and serialization to produce full-plan and revision targets. Trajectory source determines the available experience, while the personal specification determines its transformation into supervision.

\paragraph{Preference learning for contextual revisions} Noura-DPO (self) starts from Noura-SFT (self) and uses self-generated alternatives sharing a revision context, day, and changed-meal scope. The construction evaluator assesses each alternative after substitution into the surrounding plan. An alternative credited with reducing findings supplies the preferred executed response. Alternatives introducing findings, followed by those leaving findings unchanged, supply rejected responses. Matching the scope anchors the comparison to the same revision decision. We optimize the DPO loss \cite{rafailov2023dpo} against a frozen reference equal to the starting SFT checkpoint, together with a chosen-response negative-log-likelihood term. The specification supplies the preference labels, while the reference model supplies the likelihood ratios.

\paragraph{Reward learning for initial plans} Noura-GRPO (self) extends the self-generated training route with group-relative optimization \cite{shao2024deepseekmath}, initialized from the DPO checkpoint. The specification scores fresh, grounded plans at their model-written quantities, rewarding fewer clinical, plausibility, cross-day, and grounding findings before portion optimization or iterative repair. Appendix~\ref{sec:method_details:6} gives the reward, sampling, and optimization settings.

We use low-rank adaptation \cite{hu2022lora}, merge adapters for evaluation, and maintain matching training and serving templates. All reported planners are evaluated through the same execution interface. Training participants are disjoint from development and test participants. Appendix~\ref{app:method_details} gives corpus sizes, objective coefficients, construction versions, optimization settings, and checkpoint lineage.

\section{Experiments}
\label{sec:experiments}

We address four questions through population evaluation and controlled comparisons:

\noindent\textbf{RQ1: Reliability.} Can Noura produce reliable plans across a multimorbid population (Section~\ref{sec:results:reliability})?

\noindent\textbf{RQ2: Execution-based supervision.} How do executed targets and specification-guided target construction contribute to learning (Section~\ref{sec:results:supervision})?

\noindent\textbf{RQ3: Quality and efficiency.} How do training objectives and model backbones affect planning quality and efficiency (Section~\ref{sec:results:posttraining})?

\noindent\textbf{RQ4: Personalization and robustness.} How specific are plans to their recipients, how stable are training gains, and how does post-training affect external capabilities (Section~\ref{sec:results:personalization})?

\paragraph{Participants and tasks}
Tasks combine NHANES health information and dietary recalls \cite{cdc_nhanes_dietary} with FNDDS food composition \cite{usda_fndds}, requesting a seven-day plan from each participant's health context and observed intake. The population cohort contains 4,993 held-out participants with at least two of hypertension, diabetes, cardiovascular disease, and a laboratory-defined CKD phenotype. Controlled analyses use a separate 300-person test, including a fixed 150-person API-comparison subset. Qwen3-14B and GPT-6 Luna are evaluated on the full population. Cross-family replication uses a fixed 1,249-person subset. Configuration development uses another 50 participants. Cohort definitions and output completeness appear in Appendices~\ref{app:evaluation_protocol} and~\ref{app:population_transfer}.

\paragraph{Comparisons and training}
We compare direct generation, which grounds foods without optimizing portions; model-directed tool use, where the model chooses when to retrieve, check, and solve portions; and specification-guided execution (Section~\ref{sec:method:2}). Qwen3-14B is the principal platform, with Qwen3-8B and Qwen3-32B providing scale comparisons. Self-training is evaluated on Qwen3-14B, Qwen3-8B, Phi-4, and Llama-3.1-8B. The self-gen\-erated corpora contain up to 1,500 trajectories from training participants disjoint from evaluation. Noura-SFT additionally uses GPT-6 Luna trajectories on the same participants. All systems use the same frozen evaluation specification and fixed participant manifests. The principal Noura-SFT checkpoint must meet a prespecified HEI-2020 criterion on the controlled set before assessment on the disjoint population cohort. Appendices~\ref{app:method_details} and~\ref{app:evaluation_protocol} give training and execution settings.

\paragraph{Metrics}
The primary endpoints are complete-plan pass rate (CPR), the percentage of participants whose plan has no evaluable clinical, plausibility, cross-day, or grounding failure, and rule violations (RV), the number of failed rule--participant--day checks. Rule-violation rate (RVR) divides RV by applicable rule-day checks. Daily pass rate (DPR) is the percentage of passing days, excluding cross-day checks. Critical-violation prevalence (CVP) is the percentage of participants with any critical failure. HEI-2020 measures dietary quality by pooling each participant's seven-day intake and averaging participant scores. HEI reflects alignment with general dietary guidelines and does not assess individualized dietary requirements \cite{nci_hei_uses}. We therefore interpret it alongside specification satisfaction and habit fit rather than as a standalone measure of clinical appropriateness. Habit fit is the share of executed energy from observed foods or suggested substitutes. Efficiency uses output tokens across all planning calls and amortized batch time (ABT), batch wall time divided by participant count. Paired intervals resample participants; denominator, unresolved-check, and bootstrap conventions appear in Appendix~\ref{app:evaluation_protocol}. External evaluation reports position-debiased QA accuracy and strict prompt-level IFEval accuracy (Appendix~\ref{app:transfer_robustness}).

\section{Results}
\label{sec:results}

\subsection{Reliable planning at population scale (RQ1)}
\label{sec:results:reliability}

Post-training improves complete-plan feasibility within a fixed execution protocol. Using Qwen3-14B and GPT-6 Luna trajectories, Noura-SFT raises complete-plan passing from 24.0\% to 72.5\% on 4,993 held-out participants (Table~\ref{tab:population_performance}). The execution protocol is unchanged, so this comparison measures what post-training adds to a planner already supported by grounding, optimization, and revision. Rule failures decrease and HEI increases, with no observed critical failures. Gains extend across four condition domains and three multimorbidity strata (Appendix~\ref{app:disease_subgroups}).

\begin{table}[t]
\centering
\caption{Population evaluation on 4,993 held-out participants. CPR, DPR, and CVP are percentages; RV counts failed rule-day checks; HEI is mean HEI-2020. Bold marks the best Qwen3-14B values; -- denotes unevaluated HEI.}
\label{tab:population_performance}
\footnotesize
\setlength{\tabcolsep}{2.4pt}
\begin{tabular}{@{}lrrrrr@{}}
\toprule
Configuration & CPR $\uparrow$ & DPR $\uparrow$ & RV $\downarrow$ & CVP $\downarrow$ & HEI $\uparrow$ \\
\midrule
\multicolumn{6}{@{}l}{\textit{Qwen3-14B}} \\
Direct generation & 0.5 & 18.6 & 40,076 & 8.0 & -- \\
Base + execution & 24.0 & 90.0 & 1,587 & 0.0 & 80.0 \\
Noura-SFT (self) & 41.0 & 95.3 & 641 & 0.0 & 77.4 \\
Noura-DPO (self) & 50.6 & 95.9 & 547 & 0.0 & 75.2 \\
Noura-SFT & \textbf{72.5} & \textbf{99.1} & \textbf{39} & 0.0 & \textbf{83.4} \\
\midrule
\multicolumn{6}{@{}l}{\textit{GPT-6 Luna reference}} \\
Direct generation & 1.9 & 31.0 & 28,007 & 6.2 & -- \\
With execution & 82.6 & 98.9 & 91 & 0.0 & 84.1 \\
\bottomrule
\end{tabular}
\end{table}

Learning from self-generated experience supplies a teacher-free route to better population planning. Both SFT and DPO improve complete-plan passing and reduce rule failures, although HEI is lower than for the base planner (Section~\ref{sec:results:posttraining}). The principal Noura-SFT configuration also learns from teacher experience. Its 14B planner incurs fewer rule failures than GPT-6 Luna (39 versus 91), while Luna retains higher complete-plan passing. This ordering separates clinical-rule compliance from full weekly feasibility, which additionally requires coherent food choices across days (Appendix~\ref{app:api_models}).

The untrained planners' results clarify what this learning builds on. For the base planner, execution reduces the number of participants with critical failures from 399 to zero in the population cohort. Even GPT-6.1 Sol passes at most 12 of 150 controlled plans with explicit rules, higher reasoning effort, or model-directed tools, compared with 136 under specification-guided execution (Table~\ref{tab:diagnostic}). Making requirements available leaves a substantial gap that evaluating and revising concrete plans can close.

The component controls connect this gap to two complementary operations. Portion optimization adjusts quantities within a food set, while guided revision changes food choices and their arrangement across days. Removing the former raises rule failures 5.9-fold. Removing edit suggestions instead sharply increases cross-day findings. Retrieval introduces another trade-off, improving HEI by 3.2 points (95\% CI $[1.6,4.8]$) while lowering complete-plan passing (Appendix~\ref{app:execution_controls}).

\subsection{Learning from executed decisions (RQ2)}
\label{sec:results:supervision}

Execution changes the quantitative decisions the policy learns to imitate. Both formats in Table~\ref{tab:target_control}A specify food identities and quantities, but executed targets incorporate portion changes made under the personal specification. With shared prompts and admission, and matched updates, they reduce rule failures (paired difference $-88$, 95\% CI $[-132,-49]$) and raise HEI by 1.3 points ($[0.2,2.5]$). The benefit is clearest in rule compliance and dietary quality, while the complete-plan gain remains uncertain ($+7$, $[-9,23]$).

\begin{table}[t]
\centering
\caption{Supervision controls for Qwen3-14B on 300 test participants. A shares trajectories, admission with complete revision credit, and 3,631 prompts. B uses the same GPT-6 Luna trajectories from 1,500 training participants. Rows and updates report filtered training examples and optimizer steps, respectively. CPR/DPR are percentages.}
\label{tab:target_control}
\footnotesize
\setlength{\tabcolsep}{2.5pt}
\begin{tabular}{@{}lrrrrrr@{}}
\toprule
Target & Rows & Updates & CPR $\uparrow$ & RV $\downarrow$ & DPR $\uparrow$ & HEI $\uparrow$ \\
\midrule
\multicolumn{7}{@{}l}{\textit{A. Self-generated: target content}} \\
Model-written & 3,631 & 456 & 28.3 & 213 & 85.5 & 76.8 \\
Executed & 3,641 & 456 & 30.7 & 125 & 90.4 & 78.1 \\
\midrule
\multicolumn{7}{@{}l}{\textit{B. Teacher-generated: full construction}} \\
Initial drafts & 1,500 & 188 & 21.3 & 135 & 76.7 & 82.9 \\
Executed + revisions & 3,563 & 446 & 68.7 & 14 & 98.5 & 81.4 \\
\bottomrule
\end{tabular}
\end{table}

Teacher experience also benefits from this construction. Using GPT-6 Luna trajectories from the same 1,500 participants, training on executed plans and revisions raises complete-plan passing from 21.3\% to 68.7\% relative to imitating initial drafts, with fewer rule failures but lower HEI (Table~\ref{tab:target_control}B). The constructed examples expose the student to quantitative corrections and contextual repairs developed during planning. This comparison evaluates the full procedure, jointly changing target transformation, selection, revision examples, and training budget.

\begin{figure}[t]
\centering
\includegraphics[width=\columnwidth,trim=0bp 7bp 0bp 4bp,clip]{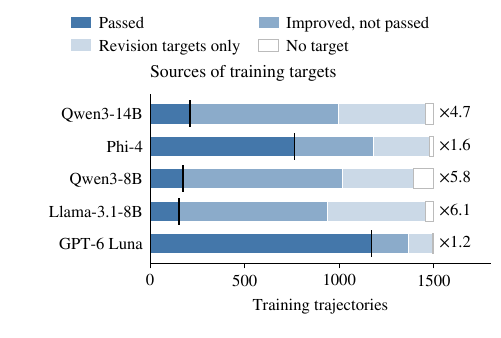}
\caption{Supervision beyond complete success. Training participants are grouped by admitted target type, from up to 1,500 trajectories per model (Phi-4: 1,499). Marks indicate the success-only pool, and multipliers compare full-plan target coverage with that pool. Corpus construction is detailed in Appendix~\ref{app:method_details}.}
\Description{Stacked bars show training participants contributing passing full-plan targets, improved full-plan targets, revision targets only, or no target across five source models.}
\label{fig:supervision_sources}
\end{figure}

Useful training decisions can be recovered before a trajectory achieves complete success. Admitting strictly improved plans expands full-plan targets from 210 to 994 among 1,500 Qwen3-14B trajectories before length filtering (Figure~\ref{fig:supervision_sources}). This broadens the experience available for self-training when fully passing plans are scarce. Appendix~\ref{app:construction_variants} examines the resulting trade-offs in feasibility, dietary quality, and cost.

\subsection{Planning quality and efficiency (RQ3)}
\label{sec:results:posttraining}

\begin{figure}[t]
\centering
\includegraphics[width=\columnwidth,trim=0bp 7bp 0bp 13bp,clip]{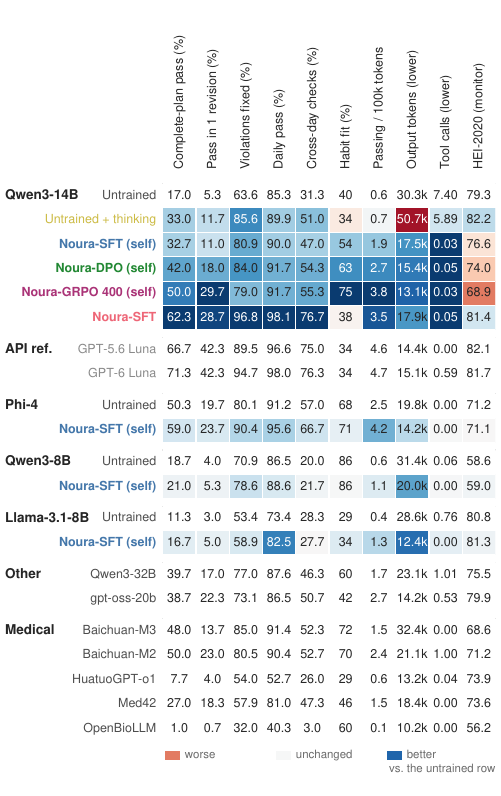}
\caption{Planning profiles on 300 participants. Cells show absolute values (k: thousands); shading shows within-column normalized changes from each backbone (blue: favorable; red: unfavorable). Metric definitions appear in Appendix~\ref{app:planner_profiles}.}
\Description{Numerical profiles of 21 planning configurations show ten metrics for outcomes, revisions, generation cost, and dietary quality. Shading compares each trained policy with its backbone. Reference and medical models are unshaded.}
\label{fig:planner_profiles}
\end{figure}

Preference and reward learning build on the initial gains from self-SFT, but change which aspects of a plan improve. On the 300-person test, DPO adds 28 passing plans over SFT (95\% CI $[11,44]$), and 400-update GRPO reaches 50.0\% complete-plan passing (Figure~\ref{fig:planner_profiles}).

The progression in passing rates does not extend to every quality dimension. Starting from DPO, GRPO raises habit fit from 63.0\% to 75.4\% but lowers HEI from 74.0 to 68.9. Plans better retain familiar foods or substitutes while scoring lower against general dietary guidelines, under a reward that excludes HEI. With additional teacher experience, Noura-SFT combines 62.3\% complete-plan passing with HEI 81.4. The training routes therefore offer different quality trade-offs (Appendix~\ref{app:posttraining_steps}).

These gains let a compact planner reach higher feasibility at lower serving cost. The 14B Noura-SFT produces 187 passing plans versus 119 for the 32B base model, with fewer rule failures and ABT of 21.2 versus 51.3 seconds. Both use one H200 and 16 workers, so this comparison measures amortized serving time. The trained planner also outperforms the thinking-enabled 14B base (ABT 77.2 seconds) in passing plans and rule failures under the same serving setting (Figure~\ref{fig:planner_profiles}).

Earlier success in the execution loop accompanies these savings. On the 4,993-person population cohort, Noura-SFT passes 37.8\% of plans after one revision, versus 7.9\% for base. Earlier passing is consistent with learning reducing the correction needed for a new participant. Feedback from training trajectories can inform subsequent proposals, while the same grounding, portion optimization, and checking procedures remain in place to evaluate and refine each generated plan. Within this cohort's separate 24-worker serving setting, ABT falls from 27.5 to 15.0 seconds (Appendix~\ref{app:population_timing}).

Reduced model-directed tool use also accompanies the savings, but disabling retrieval alone does not reproduce the training gains. Noura-SFT (self) produces 24 more passing plans (95\% CI $[7,42]$) and 58 fewer rule failures ($[-110,-11]$) than retrieval-disabled Qwen3-14B (Table~\ref{tab:execution_controls}). Automatic grounding, portion optimization, and checking remain active in both configurations, so this comparison still evaluates planning with quantitative execution.

\begin{figure}[t]
\centering
\includegraphics[width=\columnwidth]{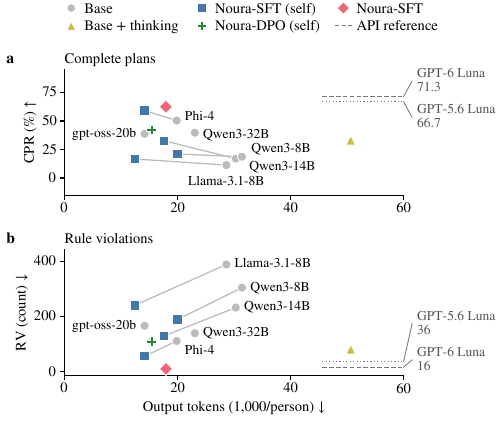}
\caption{Planning quality versus output tokens per participant on the same 300 participants. Lines link each base to its self-trained checkpoint; token comparisons are within backbone. Thinking and other trained variants use Qwen3-14B; API points provide quality references.}
\Description{Two scatter panels plot complete-plan passing and rule violations against output tokens per participant across local backbones, with API-model quality references.}
\label{fig:quality_cost}
\end{figure}

Training each backbone on its own trajectories reproduces the compliance and efficiency gains across three families, while exposing a distinction between daily and weekly feasibility. Qwen3-14B, Phi-4, Qwen3-8B, and Llama-3.1-8B all improve daily passing on the 1,249-person subset, with 41--61\% fewer rule failures and 29--55\% fewer output tokens. The controlled test shows the same pattern (Figure~\ref{fig:quality_cost}). Complete-plan passing improves for three backbones but remains largely unchanged for Qwen3-8B ($+0.9$ points, 95\% CI $[-1.9,3.7]$). Food repetition persists in 50\% of its plans and day overlap in 43\%. These weekly constraints remain obstacles to complete passing even as daily compliance improves (Appendix~\ref{app:population_transfer}).

\suppressfloats[t]

\subsection{Personal specificity and robustness (RQ4)}
\label{sec:results:personalization}

Cross-person transfer exposes two distinct demands of personalization. Of 55,913 recipient pairs formed from 187 passing Noura-SFT plans on the 300-person test, only 25 pass (0.045\%). Clinical failures occur in 59.5\% of pairs and critical failures in 20.5\%. Re-optimizing portions raises clinical compliance to 69.6\% across 5,610 sampled pairs, showing that quantities can address many clinical mismatches. Complete passing remains rare after portion adaptation. Habit-continuity failures dominate direct transfer (Appendix~\ref{app:transfer_robustness}). The available foods determine which parts of the recipient's habitual diet a menu can preserve. Food proposals therefore need to reflect the recipient's dietary context, while portion optimization aligns quantities with their clinical targets.

This continuity remains visible when plans are generated for their intended recipients. In the population cohort, Noura-SFT allocates a 2.5-percentage-point higher share of energy to observed foods or suggested substitutes than GPT-6 Luna (95\% CI $[2.2,2.9]$). Together with the lower rule-failure count, this shows that stronger clinical-rule compliance can coexist with greater dietary continuity in the learned policy.

Repeated training runs preserve the planning advantage, while external evaluations distinguish QA retention from instruction following. On the 300-person test, three Qwen3-14B self-SFT runs pass 91--98 plans versus 51 for base, and three Noura-SFT runs pass 187--197. All six reduce rule failures (seeds in Appendix~\ref{sec:method_details:3}). Across six evaluated checkpoints, position-debiased accuracy on five medical and general QA benchmarks declines by at most 1.1 percentage points. Strict IFEval changes range from $-5.7$ to $+4.4$ points, showing that instruction-following retention is less uniform than QA retention (Appendix~\ref{app:transfer_robustness}).

\section{Conclusion}
\label{sec:conclusion}

Noura connects personalized dietary planning and model learning through executable personal specifications. The same requirements coordinate food choices and quantities across a complete plan, guide revision, and provide executed targets, contextual preferences, and rewards for post-training. This construction draws supervision from useful decisions in both successful and unsuccessful trajectories, supporting learning from self-generated and stronger-model experience. A matched target-content comparison shows that executed decisions improve rule compliance and HEI-2020 relative to the corresponding model-written proposals. In an offline evaluation of 4,993 held-out NHANES participants, Noura-SFT raises the 14B planner's complete-plan pass rate from 24.0\% to 72.5\% within the same execution loop while improving HEI-2020. Self-training reduces rule violations and generated output tokens across four compact models from three families. These results show how personal requirements can serve both as criteria for execution and as supervision for learning more effective planning policies.

Future work can extend Noura to additional conditions by adding applicable rules and resolving interactions with existing requirements. For populations with different dietary practices, extensions can adapt food catalogs, habit references, and planning criteria. These extensions would test whether the same execution and learning procedure remains effective as personal requirements change.

\FloatBarrier
\label{page:main_end}
\bibliographystyle{ACM-Reference-Format}
\bibliography{refs}
\FloatBarrier
\flushbottom
\captionsetup[figure]{skip=12pt,belowskip=12pt}
\captionsetup[table]{skip=12pt,belowskip=12pt}
\appendix
\section{Implementation Details}
\setlength{\floatsep}{6pt plus 1pt minus 1pt}
\setlength{\textfloatsep}{8pt plus 2pt minus 2pt}
\label{app:method_details}

\label{sec:method_details:1}
\label{sec:method_details:2}

Controlled seven-day runs allow one draft, three repair rounds of four candidates (first greedy, others temperature 0.7), and eight information-tool calls. Information-tool and edit-suggestion interfaces use version 1 for training and version 2 for testing.

\paragraph{Independent specification review}
\label{app:specification_review}
An independent, non-author registered dietitian approved the scored rules, sources, overrides, and planning thresholds without changes, with per-rule provenance logged. Table~\ref{tab:scored_requirements} lists scored requirements. Laboratory-monitoring and medication-timing requirements remain unscored because plans cannot verify them.

\begin{table}[t]
\centering
\captionsetup{font=footnotesize,skip=2.5pt}
\caption{Scored requirements by family. Sources provide the clinical basis; numerical bounds include study operationalizations. Drug labels are FDA structured product labels from DailyMed~\cite{nlm_dailymed}; $^\dagger$critical severity.}
\label{tab:scored_requirements}
\scriptsize
\setlength{\aboverulesep}{0.35ex}
\setlength{\belowrulesep}{0.45ex}
\providecommand{\rr}{\raggedright\let\\\tabularnewline}
\setlength{\tabcolsep}{2.2pt}
\renewcommand{\arraystretch}{0.95}
\begin{tabular}{@{}p{1.10cm}p{2.35cm}p{2.65cm}p{1.90cm}@{}}
\toprule
\rr Family & \rr Applies when & \rr Implemented constraint & \rr Evidence basis \\
\midrule
\rr Sodium & \rr Hypertension; heart failure & \rr ${\le}2{,}300$; ${\le}2{,}000$ mg/day & \rr AHA/ACC hypertension~\cite{jones2025hypertension}; AHA/ACC/HFSA heart failure~\cite{heidenreich2022hf} \\
\rr Kidney & \rr CKD stage G3 or higher & \rr Phosphorus ${\le}1{,}000$ mg/day; protein ${\le}0.8$ g/kg/day & \rr KDIGO CKD-MBD~\cite{kdigo2017mbd}; KDIGO 2024~\cite{kdigo2024ckd} \\
\rr Added sugar & \rr Diabetes & \rr ${\le}10\%$ of energy & \rr ADA Standards of Care~\cite{ada2026behavior} \\
\rr Potassium, upper$^\dagger$ & \rr eGFR ${<}30$; serum K ${\ge}5.0$ mmol/L; potassium-sparing diuretics; potassium chloride & \rr ${\le}2{,}000$ mg/day (kidney, serum K); ${\le}2{,}400$ mg/day (drugs) & \rr KDIGO 2024~\cite{kdigo2024ckd}; drug labels \\
\rr Potassium, lower & \rr Loop or thiazide diuretic with serum K ${<}3.5$ mmol/L & \rr ${\ge}3{,}400$ (men), ${\ge}2{,}600$ (women) mg/day; upper bounds prevail & \rr Drug labels; NASEM~\cite{nasem2019sodium} \\
\rr Grapefruit & \rr Simvastatin, lovastatin; nifedipine & \rr No grapefruit, pomelo, or Seville orange foods & \rr Drug labels \\
\rr Tyramine$^\dagger$ & \rr MAOIs; linezolid & \rr No aged cheese, cured meat, tap beer, fermented soy, yeast extract & \rr Drug label \\
\rr Alcohol, limited & \rr Metformin; sulfonylureas; insulin & \rr ${\le}28$; ${\le}14$; ${\le}14$ g/day & \rr Drug labels \\
\rr Alcohol, none & \rr Tramadol$^\dagger$, alprazolam, acetaminophen, isosorbide mononitrate, sertraline & \rr No alcohol & \rr Drug labels \\
\rr Vitamin K & \rr Warfarin & \rr Daily within $\pm50\%$, weekly mean within $\pm20\%$ of usual intake & \rr Drug label \\
\bottomrule
\end{tabular}
\end{table}

\paragraph{Thresholds and portion solving}
Missing inputs remain unresolved. Conflicting bounds retain the upper bound and defer the lower to clinical judgment. Daily energy must lie within [0.6, 1.4] and the seven-day mean within [0.8, 1.2] of mean recalled intake (Appendix~\ref{app:energy_sensitivity}). Except for designated staples and free beverages, foods may appear on at most four days. Plans need at least 21 distinct non-free-beverage foods, day-level Jaccard similarity at most 0.8, and at least 30\% of energy from observed foods or substitutes. These bounds are operational planning criteria. Per day, the solver minimizes \(\sum_i|g_i-t_i|/\max(t_i,1)\) plus residual penalties \(\sum_j\lambda_j s_j\) subject to \(\mathbf A\mathbf g\le\mathbf b+\mathbf s\), \(\mathbf l\le\mathbf g\le\mathbf u\), and \(\mathbf s\ge0\), where \(\mathbf t\) holds proposed grams, \(\mathbf A\) encodes numerical rules, bounds are normally \([0.5,2]\,\mathbf t\), and \(\lambda_j\) ranges from \(10^5\) (critical) to \(10^2\) (minor).

\paragraph{Supervision corpora and optimization}
\label{sec:method_details:3}

Pre-filter initial-plan/post-filter training counts are 994/4,167 for Qwen3-14B, 1,183/4,097 for Phi-4, 1,015/3,568 for Qwen3-8B, and 937/3,914 for Llama-3.1-8B, from 1,500 self-generated trajectories each (Phi-4: 1,499). Noura-SFT has 2,362/7,730 targets from 3,000 Qwen3-14B and GPT-6 Luna trajectories on the same 1,500 participants. SFT uses LoRA (rank 32, \(\alpha=64\), dropout 0.05) on all attention and MLP projections, two epochs, learning rate \(10^{-4}\) with cosine decay and 3\% warmup, effective batch size 16, and bfloat16, discarding records over 16,384 tokens. Training and serving share chat templates. Completion-only loss covers initial-plan and revision targets without tool-call targets or evaluation participants. Data-order seeds are 42 (primary) and 1 and 2 (repeats).

\paragraph{Versioned revision credit}
\label{sec:method_details:4}
The Qwen3-14B, Noura-SFT, and Phi-4 corpora use original credit, zeroing candidate-side weekly energy and vitamin-K findings but retaining reference findings. Qwen3-8B and Llama-3.1-8B use complete credit, checking every finding. Retraining Qwen3-14B with complete credit gains 41 passing plans over base (95\% CI [24, 58]). Differences from original credit (10,000 resamples, seed 0) are passing plans $-6$ ([$-20$, 8]), execution-loop RV $-4$ ([$-36$, 28]), cross-day findings $+114$ ([28, 204]), and direct-generation RV $+216$ ([56, 382]).

\paragraph{Preference and reward learning}
\label{sec:method_details:6}
Within matched revision scopes (Section~\ref{sec:method:4}), the first credited alternative is preferred. Negatives prioritize added findings. DPO \cite{rafailov2023dpo} uses the frozen SFT model as reference, coefficient 0.1, and chosen-response NLL weight 1. GRPO \cite{shao2024deepseekmath} scores sampled first drafts without portion optimization or repair, with \(r=-[(R+P)/H+0.25W+0.1L+U]\): \(R\) weights daily rule findings 3/2/1/0.5 by severity; \(P\), \(W\), \(L\), and \(U\) count plausibility, cross-day, linking, and residual grounding findings; \(H\) is the horizon. Empty plans or scoring failures receive \(-10\). Each stage trains new LoRA parameters (SFT\(\to\)DPO\(\to\)GRPO, all on Qwen3-14B trajectories only). GRPO uses 400 updates over 1,600 prompts, 16 completions per update (four per prompt), learning rate $5\times10^{-6}$, temperature 0.8, a 5,120-token completion limit, group-standardized rewards, clipping 0.2, and no reference KL.

\section{Additional Experiments}
\label{app:experiments}

\subsection{Cohorts, metrics, and evaluation protocol}
\label{app:evaluation_protocol}

\paragraph{Cohorts} Eligibility requires at least four foods and 500~kcal on the first recall day. The controlled test has 300 participants selected for rule coverage. The population cohort (4,993 participants, frozen 2026-10-12) excludes training, test, and development participants. Prevalences are 89.9\% hypertension, 52.4\% diabetes, 42.9\% cardiovascular disease, and 52.3\% CKD (eGFR below 60 or albumin-to-creatinine ratio at least 30~mg/g). Participants have two (3,398), three (1,316), or four (279) domains and a median of four named prescriptions (49.3\% have five or more). The principal Noura-SFT checkpoint required a prespecified lower paired 95\% HEI bound of at least $-3$ points against base execution on the controlled test, a selection tolerance rather than clinical equivalence.

\paragraph{Models} Qwen3-14B uses its 32,768-token context~\cite{qwen2025modelcard}; Phi-4, Med42-8B, and OpenBioLLM-8B use 16,384-token contexts, with no fixed output cap for direct generation or specification-guided execution. OpenBioLLM returns 299 of 300 plans. All population runs return every requested plan. Llama-3.1-8B's principal planning outcomes use an updated self-SFT checkpoint (same corpus and hyperparameters). External tasks and the profile's violations-fixed and cross-day entries use the earlier one.

\paragraph{Metrics} CPR/CVP denominators are intended participants, and DPR uses seven days each. Missing outputs are nonpassing. Unresolved (missing inputs or deferred conflicting bounds) and advisory statuses do not block CPR, but every evaluable failure does. RVR includes unresolved and advisory rule-day checks in its denominator (13,916 controlled; 164,927 population), so it does not complement verified passing. HEI-2020 uses all 13 components~\cite{nci_hei_scoring}, averages returned plans without population weights, and is absent from target admission, preference construction, and reward.

\paragraph{Cost and uncertainty}
\label{app:population_timing}
Tokens include all planning calls and discarded candidates. Tool attempts include invalid names and exclude automatic grounding, solving, and checking. Timing uses one H200 with 16 workers (controlled) or 24 workers (population, four batches of about 1,249 participants). Paired 95\% intervals use 5,000 participant bootstrap resamples (seed 20261017).

\subsection{Planning conditions and controls}
\label{app:api_models}
\label{sec:results:diagnostic}
\label{app:execution_controls}
\label{app:execution_stages}

API runs use schema-constrained output and low reasoning effort unless marked. With model-directed tools, GPT-6.1 Sol calls the checker for only nine of 150 participants (Table~\ref{tab:diagnostic}). Critical-violation prevalence is 14.0--33.3\% under direct generation, 4.0--30.7\% with rules in the prompt, 18.7--28.0\% with model-directed tools, and at most 2.7\% under specification-guided execution. On the population cohort, Noura-SFT minus GPT-6 Luna is $-52$ rule failures ($[-85,-20]$), $-10.2$ CPR points ($[-11.5,-8.8]$), $-0.67$ HEI points ($[-0.83,-0.52]$), and $+2.52$ habit-fit points ($[2.17,2.86]$).

\begin{table}[t]
\centering
\captionsetup{font=footnotesize,skip=2.5pt}
\renewcommand{\arraystretch}{0.85}
\caption{Planning conditions on 150 participants: CPR (\%) / RV (count). Dir.: direct generation; Rules: rules in prompt; Tools: model-directed tools; Exec.: specification-guided execution; --: not evaluated. \textsuperscript{\textdaggerdbl}4,000-token output limit (6/150 replies reached it).}
\label{tab:diagnostic}
\scriptsize
\setlength{\aboverulesep}{0.35ex}
\setlength{\belowrulesep}{0.45ex}
\setlength{\tabcolsep}{2.1pt}
\begin{tabular}{@{}lrrrrrrrr@{}}
\toprule
& \multicolumn{2}{c}{Dir.} & \multicolumn{2}{c}{Rules} & \multicolumn{2}{c}{Tools} & \multicolumn{2}{c}{Exec.} \\
\cmidrule(lr){2-3}\cmidrule(lr){4-5}\cmidrule(lr){6-7}\cmidrule(l){8-9}
Model & CPR & RV & CPR & RV & CPR & RV & CPR & RV \\
\midrule
Qwen3-8B & 0.0 & 1,937 & -- & -- & -- & -- & 16.7 & 124 \\
Qwen3-14B & 0.0 & 1,960 & 0.0 & 1,938 & 0.0 & 1,693 & 18.7 & 121 \\
Qwen3-32B & 0.0 & 1,638 & -- & -- & -- & -- & 38.7 & 85 \\
Phi-4 & 0.0 & 1,470 & -- & -- & 0.0\textsuperscript{\textdaggerdbl} & 1,354\textsuperscript{\textdaggerdbl} & 48.0 & 50 \\
gpt-oss-20b & 0.0 & 1,148 & -- & -- & 0.0 & 1,147 & 37.3 & 107 \\
Llama-3.1-8B & 0.0 & 2,083 & -- & -- & -- & -- & 12.0 & 171 \\
Med42-8B & 0.0 & 1,755 & -- & -- & -- & -- & 26.7 & 153 \\
OpenBioLLM-8B & 0.0 & 1,032 & -- & -- & -- & -- & 1.3 & 324 \\
\midrule
GPT-5.6 Terra & 0.7 & 1,752 & -- & -- & -- & -- & -- & -- \\
GPT-5.6 Luna & 0.7 & 1,592 & 0.7 & 1,238 & 1.3 & 1,560 & 71.3 & 17 \\
GPT-6 Luna & 0.7 & 1,168 & 4.7 & 766 & 3.3 & 1,281 & 72.7 & 5 \\
GPT-6 Astra & 0.7 & 1,517 & -- & -- & -- & -- & -- & -- \\
GPT-6.1 Sol & 2.7 & 1,459 & 8.0 & 903 & 0.0 & 1,579 & 90.7 & 7 \\
\quad high effort & -- & -- & 7.3 & 872 & -- & -- & -- & -- \\
\bottomrule
\end{tabular}
\end{table}

Table~\ref{tab:execution_controls} compares Qwen3-14B settings. Component removals hold other settings fixed. Initial-plan optimization adds retrieval and portion solving. Self-review gets applicable rules without execution feedback. Check feedback gets verifier findings and edit suggestions with 12 lookups instead of eight.

\begin{table}[t]
\centering
\captionsetup{font=footnotesize,skip=2.5pt}
\renewcommand{\arraystretch}{0.85}
\caption{Qwen3-14B execution controls on 300 participants. CPR and CVP in \%; X-day: cross-day findings; ABT: seconds per participant; --: not reported.}
\label{tab:execution_controls}
\scriptsize
\setlength{\aboverulesep}{0.35ex}
\setlength{\belowrulesep}{0.45ex}
\setlength{\tabcolsep}{2.2pt}
\begin{tabular}{@{}lrrrrrr@{}}
\toprule
Configuration & CPR$\uparrow$ & RV$\downarrow$ & CVP$\downarrow$ & X-day$\downarrow$ & HEI$\uparrow$ & ABT$\downarrow$ \\
\midrule
Direct generation & 0.0 & 3,800 & 26.7 & 1,317 & 73.9 & 4.2 \\
Rules + self-review & 0.7 & 3,783 & 27.7 & 1,511 & -- & 20.3 \\
Check feedback & 3.0 & 1,821 & 14.3 & -- & 76.0 & -- \\
Initial-plan optimization & 1.0 & 705 & 1.3 & 2,176 & 80.1 & 7.1 \\
\midrule
Complete execution loop & 17.0 & 232 & 0.0 & 674 & 79.3 & 38.3 \\
\quad w/o portion optimization & 4.7 & 1,372 & 11.0 & 1,124 & 76.9 & 39.4 \\
\quad w/o edit suggestions & 4.7 & 412 & 0.0 & 1,939 & 79.8 & 45.3 \\
\quad one candidate per round & 13.3 & 333 & 0.3 & 860 & 80.0 & 14.9 \\
\quad w/o information retrieval & 24.7 & 187 & 0.0 & 509 & 76.1 & 27.7 \\
\bottomrule
\end{tabular}
\end{table}

\subsection{Supervision controls and post-training}
\label{app:supervision_controls}
\label{app:construction_variants}
\label{app:posttraining_steps}

In Table~\ref{tab:target_control}A, the two corpora share 3,631 prompts and differ by 0.046\% in tokens per epoch. Executed targets add 104 passing days ($[61,150]$) and lower habit fit by 3.13 points ($[-5.93,-0.47]$). In Table~\ref{tab:target_control}B, the executed-target pipeline cuts unresolved food occurrences from 944 to nine, raises habit fit by 6.67 points ($[5.20,8.14]$), and lowers HEI by 1.50 ($[-2.09,-0.91]$). Relative to the retrieval-disabled base planner (Table~\ref{tab:execution_controls}), Noura-SFT (self) changes HEI by $+0.5$ ($[-1.1,2.1]$). On the controlled test, with nearly matched updates, a simpler recipe using successful full plans and whole improving revisions yields 37 more passing plans for self-training (135 versus 98; $[19,54]$) and 41 more for mixed-source training (228 versus 187; $[24,59]$). Noura-SFT (self) keeps 2.1 points higher HEI ($[0.6,3.7]$) and 24\% fewer output tokens, while the simpler recipe has fewer rule failures (60 versus 129). For mixed-source training, the simpler recipe yields 7 versus 10 rule failures, HEI $+0.5$ ($[-0.1,1.1]$), and 9\% fewer tokens. Complete successes constitute 14\% of Qwen3-14B and 78\% of GPT-6 Luna trajectories. Both recipes retain improved revisions, so this compares construction recipes rather than testing whether unsuccessful trajectories are useful. GRPO incurs 133 rule failures. Relative to its DPO initialization, it changes HEI by $-5.10$ points ($[-6.59,-3.62]$) and habit fit by $+12.39$ ($[9.29,15.37]$). Among Qwen3-14B base and Noura planners with full execution, only Noura-SFT (self) incurs critical failures on the controlled test (0.7\%). On the population cohort, DPO adds 477 passing plans to Noura-SFT (self) ($[409,547]$), with HEI $-2.21$ ($[-2.50,-1.93]$) and habit fit $+7.85$ ($[7.21,8.51]$).

\paragraph{Planning profiles}
\label{app:planner_profiles}
Figure~\ref{fig:planner_profiles} includes five untrained medical models among 21 configurations: Baichuan-M3-235B, Baichuan-M2-32B, HuatuoGPT-o1-8B, Med42-8B, and OpenBioLLM-8B. Baichuan-M3 uses its official 4-bit weights with thinking off. About 11,000 output tokens per plan are tool-step text discarded upon reaching the output limit. Passing within one revision includes initial successes. \textit{Violations fixed} is net percentage RV reduction from optimized initial to final plans. \textit{Cross-day checks} is the percentage of plans without weekly findings other than habit fit.

\subsection{Cross-family replication and robustness}
\label{app:population_transfer}
\label{app:transfer_robustness}
\label{app:disease_subgroups}
\label{app:energy_sensitivity}

\begin{table}[t]
\centering
\captionsetup{font=footnotesize,skip=2.5pt}
\renewcommand{\arraystretch}{0.85}
\caption{Base and self-SFT on the 1,249-person subset (all complete, none critical). $\Delta$RV: paired [95\% CI]. Tokens: k per participant, within family.}
\label{tab:population_transfer}
\scriptsize
\setlength{\aboverulesep}{0.35ex}
\setlength{\belowrulesep}{0.45ex}
\setlength{\tabcolsep}{2.2pt}
\begin{tabular}{@{}lrrrrrr@{}}
\toprule
 & CPR$\uparrow$ & RV$\downarrow$ & $\Delta$RV [95\% CI] & DPR$\uparrow$ & HEI & Tok.$\downarrow$ \\
\midrule
Qwen3-14B & 23.7 & 352 & & 90.5 & 79.9 & 27.5 \\
\quad self-SFT & 40.7 & 143 & $-209\;[-282,-140]$ & 95.8 & 77.4 & 16.3 \\
Phi-4 & 59.7 & 111 & & 95.5 & 71.5 & 18.4 \\
\quad self-SFT & 71.4 & 65 & $-46\;[-77,-16]$ & 98.0 & 71.0 & 13.1 \\
Qwen3-8B & 27.5 & 569 & & 92.2 & 59.5 & 29.2 \\
\quad self-SFT & 28.4 & 223 & $-346\;[-456,-240]$ & 95.2 & 59.5 & 19.4 \\
Llama-3.1-8B & 15.5 & 712 & & 82.4 & 80.6 & 27.2 \\
\quad self-SFT & 21.0 & 380 & $-332\;[-430,-238]$ & 90.7 & 80.9 & 12.2 \\
\bottomrule
\end{tabular}
\end{table}

\begin{table}[t]
\centering
\captionsetup{font=footnotesize,skip=2.5pt}
\renewcommand{\arraystretch}{0.85}
\caption{Change from base (points): position-debiased QA, strict IFEval; MMLU-M/G: medical/general. Other rows: self-SFT (Llama: earlier checkpoint).}
\label{tab:transfer_capabilities}
\scriptsize
\setlength{\aboverulesep}{0.35ex}
\setlength{\belowrulesep}{0.45ex}
\setlength{\tabcolsep}{2.4pt}
\begin{tabular}{@{}lrrrrrr@{}}
\toprule
Model & MedQA & MedMCQA & PubMedQA & MMLU-M & MMLU-G & IFEval \\
\midrule

14B Noura-SFT (self) & $+0.2$ & $-0.5$ & $+1.2$ & $-0.2$ & $-1.0$ & $-5.7$ \\
14B Noura-DPO (self) & $+0.2$ & $-0.6$ & $+0.2$ & $0.0$ & $-1.1$ & $-5.2$ \\
14B Noura-SFT & $+0.1$ & $-0.8$ & $+0.6$ & $-0.4$ & $-0.9$ & $-3.3$ \\
\midrule
Phi-4 & $-0.3$ & $+0.4$ & $0.0$ & $-0.2$ & $+0.3$ & $+3.9$ \\
Qwen3-8B & $+0.4$ & $-1.1$ & $-0.6$ & $+0.1$ & $+0.2$ & $-1.8$ \\
Llama-3.1-8B & $+2.0$ & $+3.3$ & $+1.0$ & $+2.6$ & $+1.9$ & $+4.4$ \\
\bottomrule
\end{tabular}
\end{table}

The 1,249-person subset takes every fourth sorted participant identifier. CPR gains are $+17.0$ points for Qwen3-14B ($[14.0,20.0]$), $+11.7$ for Phi-4 ($[8.9,14.5]$), $+0.9$ for Qwen3-8B ($[-1.9,3.7]$), and $+5.5$ for Llama-3.1-8B ($[3.1,8.0]$; Table~\ref{tab:population_transfer}). Across the four condition domains and the two-, three-, and four-domain strata, passing ranges from 19.4--24.8\% for base execution to 69.4--73.7\% for Noura-SFT. Among 3,363 participants with plausible recorded energy under Goldberg cut-offs~\cite{black2000goldberg} (Mifflin--St~Jeor~\cite{mifflin1990new}, activity level 1.55), Noura-SFT improves CPR by 50.9 points ($[49.0,52.8]$), versus 48.5 in the full cohort. Removing both energy pass conditions leaves a 44.7-point gain ($[43.0,46.3]$).

Portion adaptation re-solves each source plan's candidate foods under 30 sampled recipients' specifications. Of 5,610 adapted pairs, 9 pass and 3.6\% contain a critical failure. Habit-continuity failures dominate, occurring in 55,599 served pairs. Table~\ref{tab:transfer_capabilities} compares trained and base models on MedQA~\cite{jin2020medqa}, MedMCQA~\cite{pal2022medmcqa}, PubMedQA~\cite{jin2019pubmedqa}, medical/general MMLU~\cite{hendrycks2021mmlu}, and IFEval~\cite{zhou2023ifeval}. QA accuracy aggregates log-probabilities over all cyclic option orderings. Strict prompt-level IFEval accuracy is 79.9--82.3\% for the trained Qwen3-14B checkpoints (base 85.6\%).

\end{document}